# Deep reinforcement learning with a particle dynamics environment applied to emergency evacuation of a room with obstacles

Yihao Zhang[1,2,+], Zhaojie Chai[1,+] and George Lykotrafitis[1,*]

[1]University of Connecticut, Mechanical Engineering, Storrs, CT 06269, United States

[2]Suzhou Zhito Technology Co., LTD, AI Research and Development Center, Shanghai, China

[+]Authors make equal contribution on this paper

[*]Corresponding author

**Abstract**

A very successful model for simulating emergency evacuation is the social-force model. At the heart of the model is the self-driven force that is applied to an agent and is directed towards the exit. However, it is not clear if the application of this force results in optimal evacuation, especially in complex environments with obstacles. Here, we develop a deep reinforcement learning algorithm in association with the social force model to train agents to find the fastest evacuation path. During training, we penalize every step of an agent in the room and give zero reward at the exit. We adopt the Dyna-Q learning approach. We first show that in the case of a room without obstacles the resulting self-driven force points directly towards the exit as in the social force model and that the median exit time intervals calculated using the two methods are not significantly different. Then, we investigate evacuation of a room with one obstacle and one exit. We show that our method produces similar results with the social force model when the obstacle is convex. However, in the case of concave obstacles, which sometimes can act as traps for agents governed purely by the social force model and prohibit complete room evacuation, our approach is clearly advantageous since it derives a policy that results in object avoidance and complete room evacuation without additional assumptions. We also study evacuation of a room with multiple exits. We show that agents are able to evacuate efficiently from the nearest exit through a shared network trained for a single agent. Finally, we test the robustness of the Dyna-Q learning approach in a complex environment with multiple exits and obstacles. Overall, we show that our model can efficiently simulate emergency evacuation in complex environments with multiple room exits and obstacles where it is difficult to obtain an intuitive rule for fast evacuation.

**Introduction**

Emergency room evacuation has frequently caused more bodily injuries and death than the actual source of panic. Several building evacuation models have been developed to evaluate and predict evacuation success under a wide range of conditions [1-5]. In general, the physical representation of pedestrian motion falls into two categories: Driven or Self-Driven many-particle systems [6]. A driven many-particle systems, such as flow of fluids and granular media, considers additional interaction with the environment as the source of external driving forces. On the opposite, the driving force in a self-driven many-particle system is associated with each single agent, not exerted from outside [7, 8]. One of the well-established self-driven models used in emergency room evacuation is the social-force model [9]. Also, Bottinelli et al. use a self-driven active matter model for situations where large groups of people with common interest gather together [10]. In addition, experimental studies investigate the behavior of pedestrians using recorded data. For instance, Gallup et al. investigated the influence of emotional facial expression on the walking behaviors of pedestrians [11]. In a recent work, Garcimartin et al. experimentally demonstrate that the pedestrian flow rate is not necessarily altered by the presence of an obstacle [12]. An important aspect of understanding emergency room evacuation is the development of models that can predict the emergent collective crowd behavior as a result of individual motion and identify parameters determining evacuation efficiency and safety.

It has been suggested that the motion of people in emergency can be described as if they are subjected to "social forces" in combination with pure mechanical environmental forces. The origin of the social forces is considered to be a combination of the internal motivation of an agent to rush to an exit and the psychological tendency to avoid other agents. However, the social force model does not consistently derive optimal agent trajectories in complex environments [13]. In these cases additional rules, stemming from behavioral heuristics, are introduced to the model. One such approach adapts walking speeds and directions based on visual information related to the existence of obstacles in lines of sight [13]. In another approach, the introduction of a tactical layer that controls the choice of desired directions in combination with an operational layer that controls interactions between pedestrians and the environment can significantly improve the behavior of the social force model in complex environments [14, 15]. In dynamic environments multi-layered methods combined with fast marching methods can generate optimal trajectories [16].

Here, we propose the introduction of a deep reinforcement learning (DRL) algorithm which can determine the most efficient evacuation policy based on the cumulative reward that an agent receives from the environment. The proposed method can be seen as a type of a tactical layer that can determine the best route-choice leading to an exit avoiding obstacles [14]. DRL is an exciting subject of artificial intelligence that has recently achieved a great success in many areas including control, robotics, and games [17-22]. The general learning setting of reinforcement learning (RL) is the following: an agent resides at a state $s_t$, performs an action $a_t$ that results in a new state $s_{t+1}$ and receives a reward $R_{t+1}$. In RL, agent learns the optimal policy by discovering the trajectory associated with the maximum cumulative reward via state transition and interaction with the environment during training [23]. Several related studies using traditional reinforcement learning techniques on evacuation problems are worth mentioning. Wharton simulated building evacuation problems using a multi-agent reinforcement learning approach [24]. The building is represented by a grid-like space with rooms and fire. The state-action matrix structure is used to denote the action values for each grid position. It is shown that the multi-agent model works in large space with fire spreading uncertainly. However, this simplified grid-based method is not able to generate an accurate representation of a realistic evacuation scenario as the discrete finite grids are not sufficient to simulate agent motion and collision due to the sequential nature of this method. In another case, Tian and Jiang applied a reinforcement learning model to evacuation of a tunnel in fire [25]. Important parameters of personnel escape are obtained from numerical simulation and actual field experiments. In addition, they analyzed personnel escape and developed a mathematical algorithm for minimum evacuation time. However, a major drawback of the traditional reinforcement learning method is that it is not efficient to generate solutions in high dimensional and partially observable environments.

Until very recently, the immense dimension of the configuration space, "curse of dimensionality", prohibited the application of RL methods in the study of agent motion. Introduction of deep neural networks (DNN), in combination with powerful computational techniques and advances in computer hardware, made the application of RL methods in solving complicated problems of agent motion in unknown environments possible [19-21]. The RL algorithm that incorporates DNN is referred as DRL and it has emerged as a very powerful methodology in the study of agent motion in unknown environments. Within this framework, application of DRL in emergency evacuation problems is very promising. In a recent work, Sharma et al. used deep Q-Learning with Q-Matrix transfer learning in a graph based fire evacuation environment, which includes features like bottlenecks, fire spread, and uncertainty in performed actions [26]. They showed that the method outperforms some state-of-the-art reinforcement learning algorithms, such as Deep Q-Networks (DQN), advantage-actor-critic (A2C), and SARSA. The model also works well in a large real world fire evacuation emergency. However, this graph based model is not able to consider a realistic continuous motion of agents since it does not represent the locations of the agents in a continuous space explicitly. Additionally, room evacuation often occurs in a multiple agents system where a number of agents trying to find the shortest time to escape in a collective manner. A big challenge for DRL applied on multiple agents system is that agents not only interact with the environment but also with each other. In such cases, the environment is non-stationary as all agents potentially interact with each other and learn concurrently. As a consequence, the convergence theory of Q-learning applied in single agent learning is not guaranteed to hold in a non-stationary environment. Here, we address this challenge by employing a single agent transfer learning approach using a shared neural network. As we will show in the results, the network learned by a single agent can be transferred to other agents in the case of multiple agent evacuation resulting in efficient collective motion.

In this paper, we develop a room evacuation DRL computational model employing a particle dynamics environment. By 'emergency' we mean that agents must evacuate the room as soon as possible. In DRL this results in the necessity of identifying the policy of fastest evacuation. Agents, exits, walls, and obstacles are explicitly represented in the simulation environment. The interaction between agents and environment is described by a ground physics rule, which includes a self-driven force, an avoidance force, a physical compression force, and a friction force between agents and between agents and obstacles, as well as a viscous damping force determining how fast the actual velocity is adjusted to the desired velocity during evacuation. DRL is used to predict the most efficient emergent motion of agents during evacuation. We approximate the action value functions by a neural network comprising multiple perceptrons. The state of an agent is described by its position and its velocity. The output is 8 discrete action values related to 8 directions available to the self-driven force. Then, we apply Dyna-Q, a RL approach that integrates both model-free direct RL using real experience and model-based planning from memory replay, to obtain the action value functions and improve the evacuation policy. During the training session, an agent learns to evacuate the room in the shortest time by updating its network implementing the $\varepsilon$ greedy exploration-exploitation technique, and experience replay with train-target network update. Upon training completion, the trained network is transferred to other agents evacuating the same room. As we increase the numbers of doors and introduce obstacles, we follow the same procedure as the one briefly described above to train agents to efficiently evacuate the room. In those cases however, we adjust the number of hidden layers and the number of neurons of the neural network.

**Models and Methods**

In this section we describe the modeling methodology. First, we describe the simulation, or "gaming", environment. It is common in modern DRL applications that the input data are extracted from a series of images as the game unfolds. An important information extracted by these series of images is the "gaming" environment which is then fed to the neural network. In other words, the recorded data provide the rules of the underlying physics that governs the motion of the agents. For example in the case of the video game *Pong*, the series of images provide the underlying physics for the motion of the ball and the collisions between the ball and the wall or the ball and the paddle. We note however that the "gaming" environment itself does not provide the best policy for winning the game and a player needs to train in order to learn to achieve the best performance within the constraints defined by the underlying physics. This means that the neural network extracts from the series of images the physics of how the "world" functions but not the gaming model. In this work, we use ground physics rules ("gaming" environment) for the agent motion. However, instead of inferring them from recorded images, we directly implement them via a particle dynamics model.

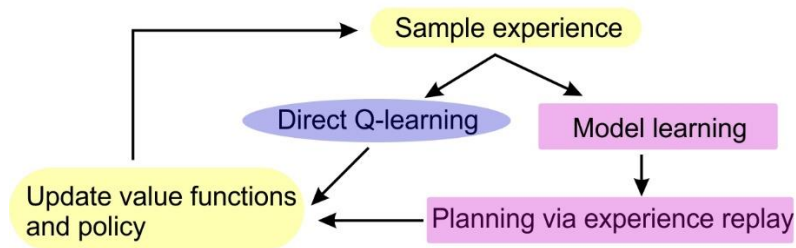

**Figure 1.** Dyna-Q training-learning schema. During a training session, the sample experience are generated following the current policy. The left loop (yellow and blue sections) means the model-free Q-learning algorithm directly updates the value functions and policy through experience. The right loop (yellow and pink sections) represents the model-based RL method via model learning and experience replay.

The DRL model is inferred via agent training within the physics environment. The training-learning schema is shown in Fig. 1. The purpose of DRL in efficient room evacuation is to train agents to evacuate as fast as possible. We consider the Dyna-Q learning algorithm for policy improvement through samples of "actual" experience. It combines the Q-learning algorithm, which is a model-free RL method, and a model-based RL method. The Q-learning is a model-free method since an agent learns directly from experience without need of prior knowledge of the model. In the case of the model-based RL method, an agent acts according to a model derived from previous experience. In this way, we generate additional data to further improve the learned action value functions and policy. A big advantage of model-based RL is that we are often able to make full use of a limited amount of experience and thus achieve a better policy.

**Particle dynamics room evacuation environment**

The simulation "gaming" environment is developed by implementing a continuous particle dynamics approach. The room is considered as a two dimensional (2D) square that contains agents, exits, walls, and obstacles. An example is shown in Fig. 2. A single agent is represented by a circle with a diameter of 0.5 m. The size of a door exit is 1.0 m and its effective region is also represented by a sphere, which means if the distance between the center of an agent and the door is less than $(0.5 + 1.0)/2 = 0.75$ m, then we consider that the agent evacuates the room successfully. The walls are the outer boundaries of the room and the obstacles are located within the room area. As agents move, the ground physics of simulation follows closely the social force model introduced by Helbing et al. [9, 27].

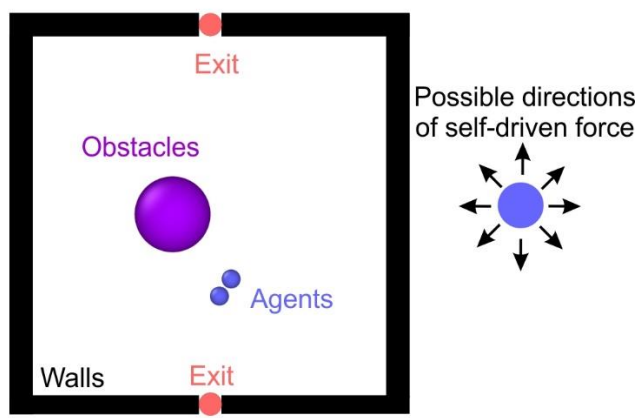

**Figure 2.** Illustration of the particle dynamics model environment. Agents, exits and obstacles are represented by circles with different colors and sizes. The walls are the outer boundaries of the room. The arrows represent the 8 possible directions of the self-driven force employed in the DRL algorithms.

(i) Self-driven force

The self-driven force represents the driving force applied to an agent to reach its desired speed and as a result it accelerates towards the exit changing its position accordingly. In the framework of the social force model, the direction of a self-driven force varies continuously. Here, we allow the self-driven force to vary in 8 discrete directions as shown in Fig. 2. The self-driven force is given by the expression:

$$\mathbf{F}_{\text{self-driven}} = \frac{m}{\tau} v_{\text{desired}} \mathbf{f}_{\text{action}}, \tag{1}$$

where m is the mass, $\tau = 0.5$ s is the relaxation time corresponding to the viscous force described later, $v_{\text{desired}}$ is the desired speed, and $\mathbf{f}_{\text{action}}$ is the unit vector representing the 8 possible discrete directions of the self-driven force. We note that the discrete self-driven force is able to reproduce a smooth motion of agents with appropriate selection of time step as shown in the results section. In addition, this simplification is compatible with the finite output of DNNs. The self-driven force directions are also described as the actions available to an agent.

(ii) Agent-agent and agent-obstacle interaction force

Agents interact with each other and with walls and other obstacles when they approach each other or when they approach obstacles closer than a specific distance. Here, we apply an exponential repulsive force to simulate the tendency of pedestrians to avoid each other. This force is given by the expression $\mathbf{F}_{ij}^{\text{avoidance}} = A \cdot \exp\left(\frac{d_{ij}-r_{ij}}{B}\right) \hat{\mathbf{r}}_{ij}$, where $A = 100$ N, $B = 0.08$ m, $d_{ij} = (d_i + d_j)/2$ is the equilibrium distance between two agents i and j, $r_{ij} = |r_i - r_j|$ is the actual distance between two agents, and $\hat{\mathbf{r}}_{ij} = \frac{r_i - r_j}{|r_i - r_j|}$ is the unit vector pointing from j to i. In addition, we implement a repulsive force $\mathbf{F}_{ij}^{\text{compression}} = kg(r_{ij} - d_{ij})\hat{\mathbf{r}}_{ij}$, which counteracts body compression, where $g(x) = x\,H(x)$ and $H(x)$ is the Heaviside function, and $k = 8.0 \times 10^4 kg/s^2$. We also implement a "sliding friction force" $\mathbf{F}_{ij}^{\text{friction}} = kg(r_{ij} - d_{ij})\mathbf{u}_{ij}$, which represents tangential interaction between pedestrians moving with different velocities. $\mathbf{u}_{ij} = (\mathbf{v}_j - \mathbf{v}_i) \cdot \hat{\mathbf{t}}_{ij}$ is the relative tangential velocity with $\hat{\mathbf{t}}_{ij} \cdot \hat{\mathbf{r}}_{ij} = 0$. The overall force applied between pedestrians is given by the expression

$$\mathbf{F}_{ij} = \mathbf{F}_{ij}^{\text{avoidance}} + \mathbf{F}_{ij}^{\text{compression}} + \mathbf{F}_{ij}^{\text{friction}} = A \cdot \exp\left(\frac{d_{ij}-r_{ij}}{B}\right) \hat{\mathbf{r}}_{ij} + kg(r_{ij} - d_{ij})\hat{\mathbf{r}}_{ij} + kg(r_{ij} - d_{ij})\Delta v_{ji}^{t} \hat{\mathbf{t}}_{ij} \tag{2a}$$

In the case of walls and other obstacles, an avoidance force, a compression force and a friction force are applied to an agent from the obstacle. The direction of the repulsive force is perpendicular to the contact surface pointing towards the agent and away from the obstacle. The overall force is given by the expression

$$\mathbf{F}_{iw} = \mathbf{F}_{iw}^{\text{avoidance}} + \mathbf{F}_{iw}^{\text{compression}} + \mathbf{F}_{iw}^{\text{friction}} = A \cdot \exp\left(\frac{d_{iw}-r_{iw}}{B}\right) \hat{\mathbf{r}}_{iw} + kg(r_{iw} - d_{iw})\hat{\mathbf{r}}_{iw} + kg(r_{iw} - d_{iw})(\mathbf{v}_i \cdot \mathbf{t}_{iw})\hat{\mathbf{t}}_{iw}, \tag{2b}$$

where $d_{iw}$ is the radius of the agent and $r_{iw}$ is the shortest distance between the center of the agent and the surface of the obstacle, $\hat{\mathbf{t}}_{iw} \cdot \hat{\mathbf{r}}_{iw} = 0$.

(iii) Viscous damping force

If only a constant self-driven force was applied to an agent, then the agent would accelerate continuously for a significant time interval and reach an unrealistic speed before it collides with another agent or an obstacle. Thus, in an environment without "viscosity", agents will move and collide violently. Because of this, we hinder unrealistically large velocities by introducing the viscous damping force:

$$\mathbf{F}_{\text{viscous}} = -\frac{m}{\tau}\mathbf{v}, \tag{3}$$

where $\mathbf{v}$ is the velocity of an agent and $\tau$ is a characteristic time that determines how fast the actual velocity $\mathbf{v}$ is adapted to the desired velocity.

In our simulations, the room is considered as a 10 m × 10 m square area with one or multiple exits and potentially with one or multiple obstacles (Fig. 2). Although the parameters for each agent can vary, we choose, without loss of generality, identical values for all agents in order to reduce the number of parameters. The initial speed is 0 m/s, the agent mass is m = 80 kg, and the exit door size of 1 m as suggested in published work [1, 28]. We have examined the behavior of the algorithm and the resulting directions of the self-driven force $\mathbf{F}_{\text{self−driven}}$ for different desired speeds – from 1 m/s to 7 m/s – to investigate if the level of the desired speed affects the result.

The motion of an agent follows the Newton equation:

$$m\frac{d^2\mathbf{r}}{dt^2} = \sum \mathbf{F}_i, \tag{4}$$

where $\mathbf{r}$ is the position and $\mathbf{F}_i$ is the total physical and social force acting on an agent $i$. For each time step, the position ($\mathbf{r}$), velocity ($\mathbf{v}$) and acceleration ($\mathbf{a}$) are updated using the leapfrog algorithm [29]:

$$\mathbf{v}\left(t + \frac{\Delta t}{2}\right) = \mathbf{v}(t) + \mathbf{a}(t)\frac{\Delta t}{2}, \tag{5}$$

$$\mathbf{r}(t + \Delta t) = \mathbf{r}(t) + \mathbf{v}\left(t + \frac{\Delta t}{2}\right)\Delta t, \tag{6}$$

$$\mathbf{v}(t + \Delta t) = \mathbf{v}\left(t + \frac{\Delta t}{2}\right) + \mathbf{a}(t + \Delta t)\frac{\Delta t}{2}, \tag{7}$$

where t is the current time and $\Delta t = 0.1$ s is the time step.

**Deep reinforcement learning algorithm**

The motion of agents in an emergency evacuation problem can take arbitrary long time until they eventually reach the exit. The time length is called horizon in RL. The room evacuation problem that we study here is considered as a problem with an indefinite but finite horizon. We use Dyna-Q learning, $\varepsilon$ greedy exploration-exploitation, and train-target network update to train agents to learn to evacuate efficiently.

The fundamental elements in RL are the following: state of environment ($s_t^e$), state of agent ($s_t^a$), action (a), reward ($R$), and policy ($\pi$). By state of environment we mean the information that represents the environment. It contains the positions and velocities of agents and the location of walls and other possible obstacles. The environment state is used to generate an observation and a reward that are transmitted to the agent. The state of an agent is its internal representation of the environment. An agent must be able to sense the state of the environment to some extent. For instance, in the case of a fully observable environment, the agent directly observes the environment. In this case, the agent state is equal to the environment state ($s_t^e = s_t^a$). In most cases, however, an agent only partially observes the environment.

To clarify the notation used in this work, by "state" we mean the position $(x, y)$ and velocity $(v_x, v_y)$ of an agent moving in the 2D simulation environment. Interaction between the agent and the environment is described in a sequence of discrete time steps. At time t, the agent selects an action ($a_t$) based on the state ($s_t$) and transitions to the next state ($s_{t+1}$) where it receives a reward ($R_{t+1}$). The sequence ($s_t, a_t, R_{t+1}, s_{t+1}$) continues until the end of each episode.

The expected cumulative reward for a state is called state value function $v_\pi(s)$ and is given by the Bellman equation below [23]:

$$v_\pi(s) = \sum_a \pi(a|s) \sum_{s',R} p(s', R|s, a)[R + \gamma v_\pi(s')], \tag{8}$$

where $\pi(a|s)$ is the policy representing the probability that the agent will choose an action (a) from the state (s), $p(s', R|s, a)$ is the probability that the agent receives a reward (R) and moves to a state ($s'$) from a state (s) after taking an action (a), whereas $0 \leq \gamma \leq 1$ is the discounting ratio.

The cumulative reward of a state (s) after taking an action (a) for the entire episode is called action value function $Q_\pi(s, a)$, often referred as Q-value, and defined by the corresponding Bellman equation:

$$Q_\pi(s, a) = \sum_{s',R} p(s', R|s, a)[R + \gamma v_\pi(s')]. \tag{9}$$

The goal in RL control is to identify the optimal policy for an agent at various states leading to a maximum cumulative reward. The resulting Bellman optimality equation for the optimal action value function $Q_*(s, a)$ is given by the solution of the equation:

$$Q_*(s, a) = \sum_{s',R} p(s', R|s, a)[R + \gamma \max_{a'} Q_*(s', )] \geq Q_\pi(s, a), \forall \pi. \tag{10}$$

A well-known method to solve Eqn. (10) is value iteration via dynamic programming. However, this method requires a full knowledge of the model, which is often not available. Another popular methods to obtain $Q_*(s, a)$ is the model-free Q-learning algorithm defined by Watkins et al. [30],

$$Q(s, a) = Q(s, a) + \alpha[R + \gamma \max_{a'} Q(s', a') - Q(s, a)], \tag{11}$$

where $\alpha$ is a parameter determining the updating speed. In the Q-learning algorithm, the learned Q-value directly approximates the optimal action value function by a one-step increment from the maximum $Q(s', a')$ in the next state. The converged $Q(s, a)$ is the optimal action value function $Q_*(s, a)$. The Q-learning algorithm significantly simplifies the learning process and it is a widely implemented technique in modern RL.

In DRL, we approximate the action value function by a DNN [19, 23]. The parametrized action value function $Q(s, a; w)$ takes the state (s) as input and outputs 8 action values through a DNN with trainable parameters (w). The action values obtained from the DNN represent the cumulative rewards when the self-driven force is chosen along each one of the 8 discrete directions. During a training session, an agent initially moves within the simulation environment following a random policy and generates experience that is stored in the memory. This experience is directly used for training to obtain the new action value function and improve the current policy via the Q-learning algorithm. We note that the learning approach used to represent the action value function is related to model-free RL. In addition, the stored experience is also used to improve model learning. A batch of sample experience is generated to further improve the policy via planning using the learned model. The method of model learning and planning in RL is referred as model-based RL. Here, we adopt the Dyna-Q learning method which integrates both the model-free and model-based RL with the Q-learning algorithm [23]. The detailed training procedure is shown in Fig. 3. We first train and update the neural network for the case of a single agent evacuation. Then, we transfer the learned DNN to all other agents that will respond similarly to evacuation. Once the model is trained, it outputs the Q-value for each action from an input state providing a prediction for the goodness of each state-action pair. The action with the highest Q-value determines the optimal policy.

---

**Dyna-Q learning algorithm**
---

**(a)** Initialize the environment: room size, exits and obstacles locations.
**(b)** Initialize the algorithm's parameters such as maximum training episodes, maximum steps per episode, and learning rate.
**(c)** Initialize the DNN for the agent: a training network and a target network.
**(d) FOR** each training episode:
    **FOR** each step:
        Determine exploration or exploitation based on ε-greedy.
        Take the action ($a_t$) at the current state ($s_t$) from the chosen policy.
        Obtain the reward ($R_{t+1}$) and the next state ($s_{t+1}$).
        Store the state transition sequence ($s_t$, $a_t$, $R_{t+1}$, $s_{t+1}$) into the memory.
        Direct Q-learning via the sequence.
    **END** of state or reach the maximum steps.
    When the memory reaches the maximum length:
        Select a mini batch of samples from previous experience randomly.
        Update the action value functions and current policy through experience replay.
    Update the target network from the training network.
**END** of training episode

---

**Figure 3.** Training procedure of the Dyna-Q learning algorithm.

There are several key points in a training session that we would like to mention. As an agent moves at each time step, a negative step reward -0.1 is given unless it reaches the exit, when a 0 end reward is given. The discounting ratio (γ) represents the importance of future rewards. For example, when γ = 0, the agent considers only the instant reward and its objective is to learn how to choose an action with the maximum instant reward ignoring the value of the cumulative future rewards. As γ approach 1, more future rewards are considered and the agent becomes more farsighted. We choose the discounting ratio γ = 0.999 to ensure a strong influence of future rewards. As a result, the longer an agent takes to evacuate, lower the reward is. Thus, the choice of rewards are compatible with the goal for fast evacuation since from a RL perspective the maximum cumulative reward is equivalent to the minimum number of total time steps. In RL, an agent finds the optimal policy by learning from experience. Then it is important to have sufficient exploration during the early stage of training episodes. Here, we use the $\varepsilon$ greedy exploration-exploitation technique with an exploration rate in the form of exponential decay:

$$\varepsilon = \varepsilon_{\text{low}} + (\varepsilon_{\text{high}} - \varepsilon_{\text{low}}) \times \exp(-\frac{4}{p}\frac{\text{ep}}{\text{ep\_total}}), \qquad (12)$$

where $\varepsilon_{\text{high}} = 1.0$ is the highest exploration rate, $\varepsilon_{low} = 0.1$ is the lowest exploration rate, (p) is the desired percentage of training episodes that reach the lowest exploration rate, (ep) is the current episode and ($\text{ep}_{\text{total}}$) is the total number of episodes. (p) signifies the percentage of the total number of episodes at which the exploration rate ($\varepsilon$) drops to ~12%, which is close to the asymptotic lowest exploration rate ($\varepsilon_{low}$). In this work, we choose p = 50%. For a simulation of 10,000 episodes, the exploration rate is high for the first few hundreds of episodes. This provides enough random experience for the agent to learn. Then, the exploration rate is reduced and at ~5,000 training episodes, the exploration rate reaches the minimum value of ~10% while the exploitation rate is ~90%. During the remaining training episodes, the DNN used to compute the optimal Q-values is eventually refined because an agent has higher chance to choose the learned optimal policy than to perform random exploration. An important part in DRL is the choice of loss function for the DNN used to update the optimal Q-value function. Here, the loss function is described as the mean squared temporal difference error between the Q-values in the target network and the train network derived from the Bellman optimality equation (10):

$$L(w_{train}) = \mathbb{E}\left[\left(R_{t+1} + \gamma \max_{a_{t+1}} Q(s_{t+1}, a_{t+1}; w_{target}) - Q(s_t, a_t; w_{train})\right)^2\right]. \tag{13}$$

The Q-learning algorithm updates the current action value $Q(s_t, a_t; w_{train})$ from the next state $R_{t+1} + \gamma \max_{a_{t+1}} Q(s_{t+1}, a_{t+1}; w_{target})$. We minimize the value of loss function via the stochastic gradient descend method that updates the weights. Note that the $(w_{target})$ term are the weights in the target network, which are one step behind those in the training network $(w_{train})$. This is because the target for the next state is used to refine the weights of the network at the current state. Then $(w_{target})$ are updated from $(w_{train})$ by a factor $(\mu)$:

$$w_{target} = w_{target} + \mu(w_{train} - w_{target}). \tag{14}$$

$(\mu)$ ranges from 0 to 1 signifying the rate of network update. When $\mu = 0$, the target network does not update. When $\mu = 1$, the target network updates directly to the newly trained network. We choose $\mu = 0.1$ in this work in order to alleviate the problem of instability while maintaining a relatively fast updating rate [19, 31]. Another important technique applied in the training session is the experience replay used for model-based planning. At each time step, we store the current state, action, reward, and next state $(s_t, a_t, R_{t+1}, s_{t+1})$ into a replay memory for model learning. During a training step, a random mini batch of these stored tuples is chosen to update the weights of the training network. This method avoids a strong correlation within a sequence of observations and reduces the variance between updates. We employ the Adam optimizer [32] with a learning rate of $1 \times 10^{-4}$ to minimize the loss function. The weights in the network are updated via TensorFlow 1.12.0 [33] and its reinforcement learning library TRFL 1.0.0. For the visualization of the evacuation, we use Ovito 2.9.0 [34].

**Results and discussion**

**Evacuation of a room with one exit**

We first study evacuation of a 10 m × 10 m room with one exit. The neural network has 3 hidden layers with 64, 128, 64 neurons respectively and an output layer with 8 neurons. The weights in the hidden layers are initialized with HE normalization [35] and the exponential linear unit function is used as the activation function in order to suppress the gradient vanishing issue in the neural network [36]. The position of an agent is rescaled to (-0.5, 0.5) to accelerate the learning speed. We note that normalization is used only in the DNN for the calculation of the 8 discrete outcomes of the Q-value function. When we apply the social force model to update positions and velocities, all calculations are performed in the original spatial dimensions.

As we mentioned in the methods section, we train the network to identify the optimal direction of the self-driven force $\mathbf{F}_{\text{self-driven}}$ depending on the agent's position and velocity. During the training session a single agent is moving within the room and eventually locates the exit. In the early stage of training episodes, due to the high exploration rate, an agent will be moving randomly until it reaches the exit. The maximum time steps for each episode is 10,000 allowing a sufficient exploration experience for the agent. To accelerate the learning speed, we adopt the transfer learning technique where an agent is first trained to evacuate a room with a larger size door of 2m for 1,000 training episodes. Then, the learned network is transferred to an agent that evacuates a same size room with a 1m door for 10,000 training episodes. In the case of a larger door size, the chance of the agent to evacuate is higher during random exploration because of a larger effective exit region. The resulting successful evacuation experience is crucial for the agent to learn the optimal Q-value function and improve the model for planning. The Q-learning algorithm updates the Q-value from the maximum action value of the next state. Then, if the next state is the end state, meaning that the agent successfully evacuates, the Q-value for the current state is updated by the true cumulative reward, which is the single step end reward. The Q-values of states far away from the end state are learned from the corresponding next states in a recursive manner due to the one-step nature of the Q-learning algorithm.

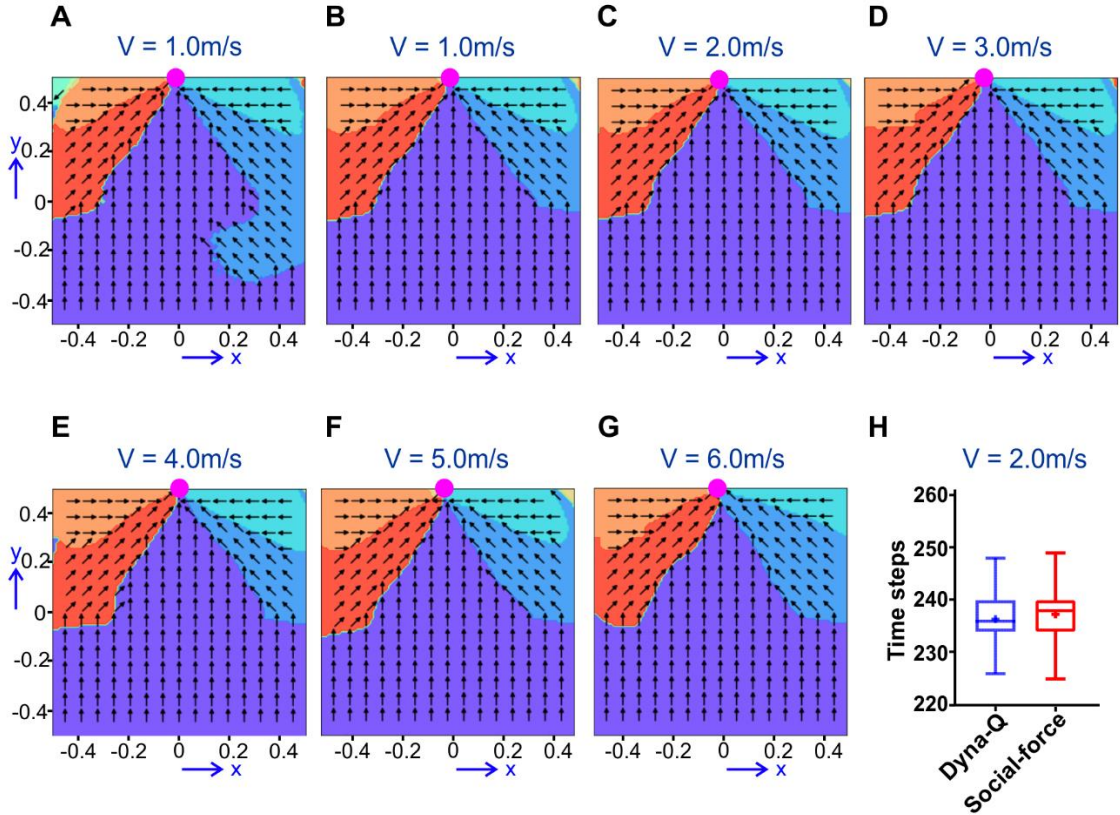

**Figure 4.** Optimal distributions of self-driven force directions in the case of zero initial speed with (A) door size of 2 m and 1 m/s desired speed after 1,000 episodes and (B) door size of 1m and 1 m/s desired speed after 10,000 episodes. (C-G) Optimal distributions of self-driven force directions in the cases of zero initial speed and 2 to 6 m/s desired speeds. The color patches and arrows show the resulting optimal distribution of the self-driven force directions. The pink circle signifies the upper exit. (H). Comparison between the number of time steps needed for complete evacuation of a 10 m × 10 m room with 80 occupants obtained using the social-force model and the Dyna-Q algorithm with zero initial speed and 2 m/s desired speed. We ran 100 cases of agents starting from random initial positions employing the Dyna-Q model and 100 cases of agents starting from different random initial positions employing the social force model. We found that there is no significant difference between the median values of the two algorithms (Mann-Whitney U test, p-value = 0.184 > 0.05).

After training is completed, the optimal policy is to choose the direction of self-driven force which has the maximum Q-value for a certain state $(x, y, v_x, v_y)$. In order to show the effect of training, we choose to illustrate the resulting distributions of the direction of the $\mathbf{F}_{self-driven}$ in the configuration space for zero initial velocities and different desired speeds (Figs. 4A to G). We find that in the case of a room with a 2 m door size and desired speed of 1 m/s and after 1,000 training episodes, the self-driven forces points to the exit, as shown by the arrows in Fig. 4A. This means that the agent is able to learn to evacuate, though it has not run enough training episodes to refine the network. When the trained agent is transferred to the case of a door size of 1 m and 1 m/s desired speed, the network is refined after 10,000 training episodes. The self-driven forces produced by the Dyna-Q approach approximately point to the exit, as shown in Fig. 4B.

According to the social-force model, the direction of the self-driven force depends on the position of the agent but not on its velocity [1]. To test the effect of the desired speed on the final distributions of the self-driven force learned by the Dyna-Q approach, we choose desired speeds of 1 m/s, 2 m/s, 3 m/s, 4 m/s, 5 m/s, and 6 m/s. The resulting self-driven force direction distributions are almost identical as shown in Figs. 4B-G respectively. We conclude that the position is the key factor to determine the direction of the self-driven force.

The trained network from one agent is used to study emergency room evacuation. To achieve this, we transfer the network trained for one agent to all agents in the room. We show that the agents evacuate the room efficiently following the policy learned from a single agent, (see Video S1). We demonstrate the behavior of 80 agents that are initially positioned in a room following a 2D uniform random distribution with zero initial speed and 2 m/s desired speed. The result shows that the agents find the exit by employing the learned policy, where we use the maximum Q-values from the output of the neural network. We note that when all agents try to evacuate from the same door, crowding and waiting occurs near the exit, which is commonly observed in a realistic situations.

Next, we test the effect of agent mass on the overall evacuation efficiency. We perform two tests within the same environment shown in Fig. 4 but with agents of different mass: 1. We consider that the mass of each agent is 80 kg; 2. We randomly assign 50% of the randomly distributed agents to have a mass of 40 kg and keep the mass of the rest at 80 kg. We set the initial speed of the agents to zero and the desired speed to 2 m/s in both cases. We run 100 cases with 80 kg mass agents distributed at random initial positions in the room and 100 cases with 50% agents of mass 40 kg and 50% agents of mass 80 kg distributed at random initial positions in the room as well. We find, by applying the non-parametric Mann-Whitney U test, that the median evacuation times are not significantly different (Fig. S1, p-value = $0.71 > 0.05$). We conclude that changes in the agents' masses, within reasonable limits, do not have a significant effect on the overall evacuation efficiency. This can be explained as follows. Eqn. (1) to (4) show that when the agent mass changes the only terms that affects the acceleration are the interaction forces between agents and between agents and obstacles, which are not critical to how fast agents reach their desired speed.

To validate the results of the Dyna-Q approach quantitatively, we compare the learned directions of the $\mathbf{F}_{\text{self-driven}}$ force to the desired directions used in the room evacuation model introduced by Helbing et al. [1] where the $\mathbf{F}_{\text{self-driven}}$ forces directly point to the exit. The efficiency of the evacuation policy learned from the Dyna-Q approach is evaluated by the number of time steps needed for all 80 agents to evacuate the room starting from random initial positions, zero initial speed, and 2 m/s desired speed. We compare the results of the two models by running 100 cases of agents starting from random initial positions for the Dyna-Q model and 100 cases of agents starting from different random initial positions for the social force model. By applying the Mann-Whitney U test we find that the median evacuation times are not significantly different (Fig. 4H, p-value = $0.18 > 0.05$). Similarly, we find that in the case of 4 m/s desired speed the median evacuation times for the Dyna-Q model and the social force model are not significantly different as well (Fig. S2, p-value = 0.30 > 0.05). In addition, we compare the results for 100 cases where the agents start each time from the same random initial positions for the Dyna-Q and the social force model while the desired speed is 2 m/s. We again find by using the Wilcoxon signed rank test that the median values are not significantly different (Fig. S3, p-value = 0.21 > 0.05). We conclude that an agent is able to learn from the Dyna-Q approach instead of following a simple rule to obtain the optimal policy for fast evacuation.

Finally, we investigate the effect of the desired speed on the overall evacuation time in the case of 80 agent room occupancy using the behavior learned from the Dyna-Q approach. The range of the desired speed is selected from 0.2 m/s to 7 m/s and the resulting number of steps required for complete evacuation is measured, as shown in Fig. S4. We observe that DRL can derive a very well-known effect in emergency room evacuation. Namely, the nonlinear dependence of evacuation time on the desired speed. As the desired speed increases to approximately 2 m/s the evacuation efficiency increases and the number of steps required for complete evacuation drops drastically. However, as the desired speed becomes larger than 2 m/s, the evacuation efficiency decreases as the number of steps required for complete evacuation increases. We note that this illustrates the "faster is slower" congestion effect as explained in [1].

**Evacuation of a room with an obstacle and one exit**

Next, we study evacuation of a room with one obstacle and one exit. First, we choose the obstacle to be of a circular shape (Fig. 5A). We consider the initial speed of the agents to be zero and the desired speed to be 2 m/s. The self-driven force distribution is similar to the case without obstacle shown in Fig. 4, except the region below the obstacle. We note that the resulting self-driven force directions around the obstacle, as illustrated in Fig. 5A, show that the DRL agent is able to learn to evacuate the room by avoiding the obstacle. Then, we compare the trajectories between the Dyna-Q algorithm and the social force model in the case of 80 agents evacuating the room, as shown in Fig. S5. The left column corresponds to Dyna-Q algorithm and the right column to the social force model. We find that the resulting trajectories are similar meaning that the evacuation efficiency of both methods are approximately the same.

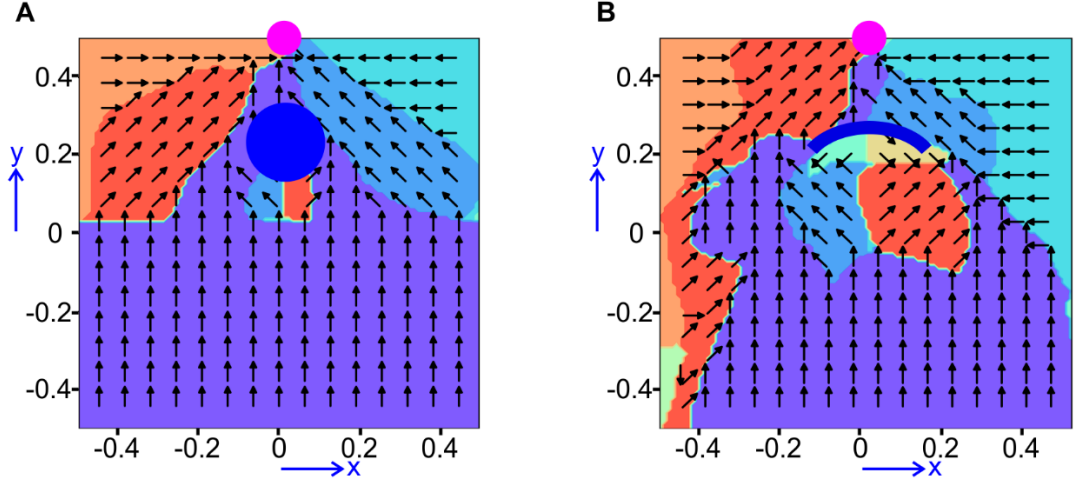

**Figure 5.** Optimal distributions of the self-driven force in the case of evacuation of rooms with (A) 1 circular obstacle and 1 exit, and with (B) 1 concave obstacle and 1 exit. The graphs correspond to the cases where the agents have zero initial speed and 2 m/s desired speed.

Second, we choose an obstacle of a concave shape in front of the exit (Fig. 5B). In this case, the Dyna-Q algorithm has a clear advantage compared to the social force model. To clearly show the different behavior of the two approaches, we compare in Fig. S6 the trajectories between the Dyna-Q algorithm and the social force model in the case of 80 agents evacuating a room with the concave obstacle. The left column corresponds to Dyna-Q algorithm and the right column to the social force model. We find that agents following the social force model can be trapped by the concave obstacle causing the unnatural result of non-complete or very delayed room evacuation (Video S2). In the DRL approach on the other hand, agents are able to identify an efficient room evacuation policy by learning to avoid the obstacle and successfully perform room evacuation (Video S3).

**Evacuation of a room with multiple exits**

Here, we study emergency evacuation of a room with 2 and 4 exits following the same training approach as in the case of one exit. Specifically, in the case of 2 exits we use the same neural network structure as in the case of one exit and train it for a 2 m door size with 1,000 episodes. Then, we transfer the network parameters and refine them for the case of a 1m door size for 10,000 episodes. We consider the initial speed of the agents to be zero and the desired speed to be 2 m/s. The results are shown in Fig. 6A and Video S4. The self-driven force distribution reveals that the initial force direction approximately points to the nearest exit. As the agent moves, the self-driven force directs the agent to follow the shortest path to the exit meaning that a trained agent is able to find the nearest exit efficiently.

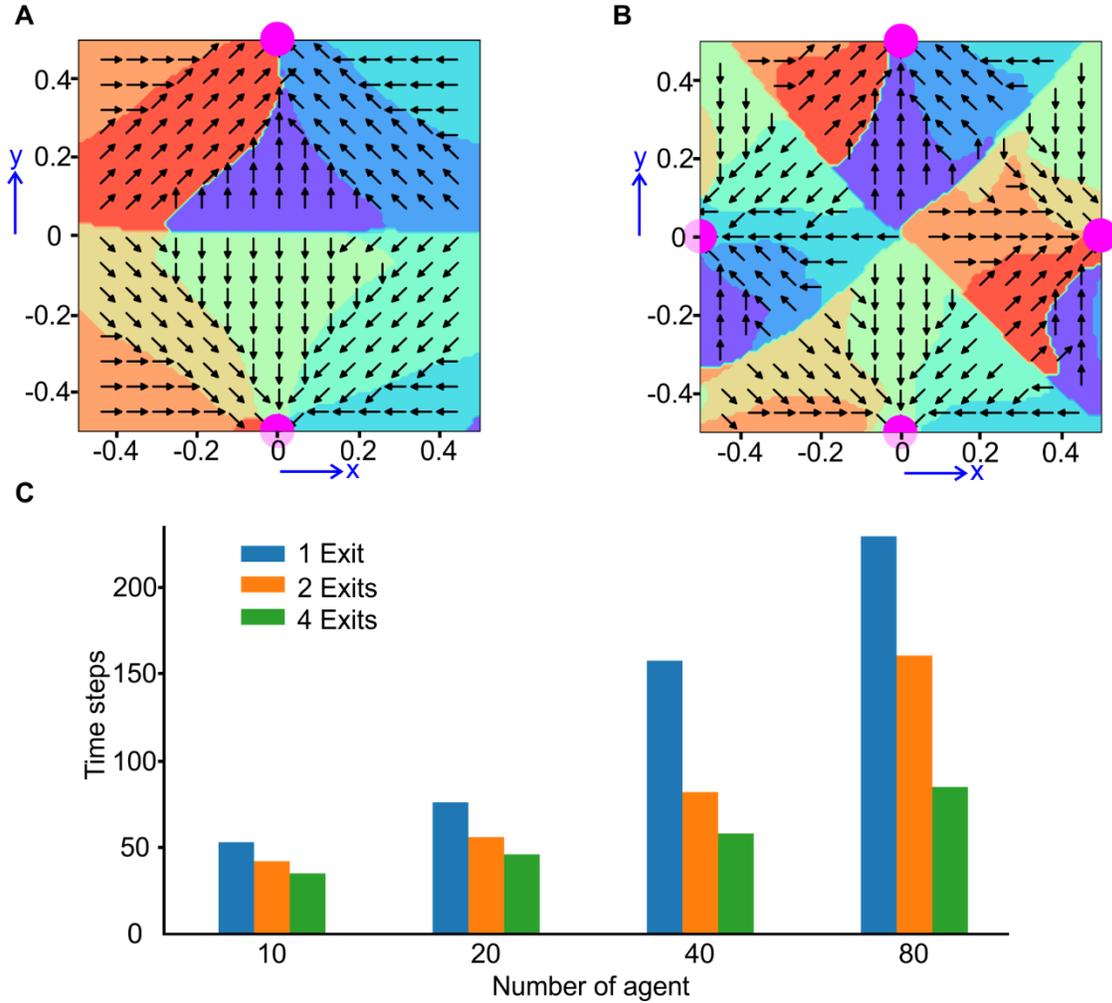

**Figure 6.** Optimal distributions of the self-driven force in the case of evacuation of rooms with (A) 2 exits and (B) 4 exits but without obstacles. In both cases clear boundaries are formed in agreement with the symmetries of the configuration spaces. The graphs correspond to the cases where the agents have zero initial speed and 2 m/s desired speed. (C) Comparison between the average total numbers of evacuation time steps for 1, 2 and 4 exits with 10, 20, 40, and 80 agents respectively. In all cases the agents have zero initial speed and 2 m/s desired speed.

Next, we increase the depth of the DNN from 3 to 4 hidden layers with 64, 128, 128, and 64 neurons respectively and an output layer with 8 neurons for the case of 4 exits. The weights in the hidden layers are again initialized with HE normalization and the activation function is the exponential linear unit function. We plot the direction of the self-driven force for the case of zero initial speed and 2 m/s desired speed. The geometric distribution of the environment in this case indicates that the boundaries along the diagonals can be used to separate areas leading to the corresponding nearest exits. Since we have shown that the position is the key factor to determine the self-driven force direction, we expect the solution to be symmetric because of the symmetric distribution of the exits. As illustrated in Fig 6B, a clear boundary appears along the diagonals of the configuration space and the self-driven force distribution is approximately symmetric. This confirms that the trained network is able to guide the agents to evacuate from the nearest exit, as shown in Video S5 as well. Additionally, we investigate how the number of doors and number of agents affect evacuation time using the learned neural networks. The total number of time steps in each case is calculated as the average value for 10,000 episodes with agents randomly distributed in the room without overlapping at the initial configuration. The results show that as the number of agent increases, the time step required for the agents to advance to the exit also increases. However, the evacuation time for the same number of agents is reduced when the number of doors increases (Fig. 6C).

**Evacuation of a room with obstacles and two doors**

First, we consider one circular obstacle located at the center of a room with two exits (Fig. 7A). The obstacle is represented as a circle with a 2 m diameter. We consider the initial speed of the agents to be zero and the desired speed to be 2 m/s. The result in Fig. 7A shows that the self-driven force distribution for stationary agents is very similar with the case without the obstacle (Fig. 6A and Video S6). This is because the obstacle does not inhibit the optimal motion since it is located at the center of the room. Next, we move the obstacle along the vertical central axis closer to the upper exit at a 2m distance from the center of the room Fig. 7B.

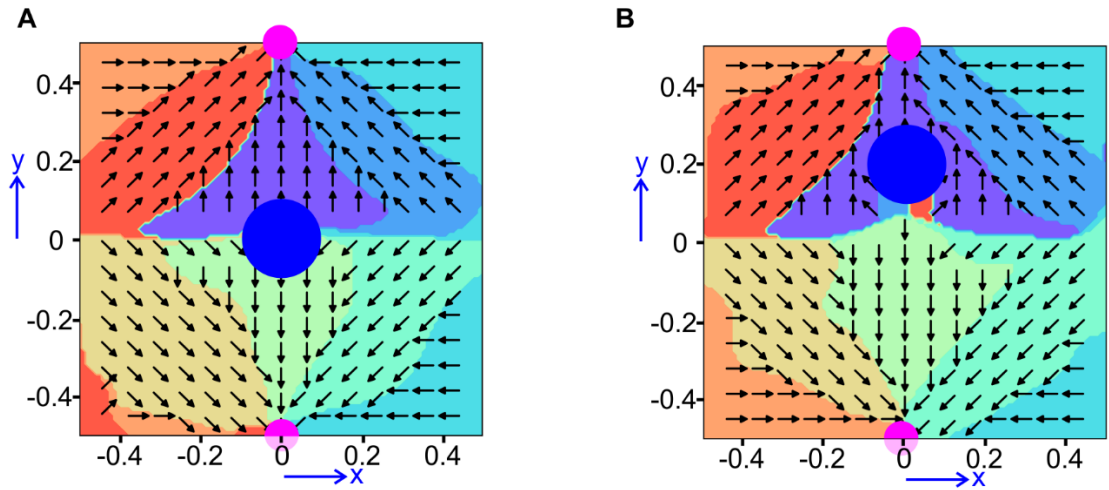

**Figure 7.** Optimal distribution of the self-driven force in the cases of (A) one round obstacle (blue circle) in the middle of a room with two exits and (B) one round obstacle closer to the upper exit door.

We find that the distribution of the self-driven forces in the area below the horizontal axis is similar with the case of two exits (Fig. 6A), as there is no obstacle within this region. However, the upper part of the self-driven force distribution map shows a different distribution due to the existence of the obstacle. It is shown that the boundary separating the general direction of evacuation is not a straight line along the horizontal direction as previously shown in Fig 6A, but curved upwards at the center of the room (Fig. 7B and Video S7). This means that an agent starting from those areas will evacuate from the lower exit because the obstacle blocks the path to the upper exit and consequently it takes longer time to go around the obstacle than heading to the lower exit. However, we found that evacuation of agents starting from other areas below the obstacle can occur from the upper exit since it was proven faster for the agents to navigate around the obstacle than moving directly to the lower exit. We note that as the obstacle comes closer to each other or to the walls, the available pathway for agents to evacuate is narrower. In such cases, the bottleneck effect becomes stronger and the narrow pathway slows down the overall evacuation since it makes impossible for the agents to pass through.

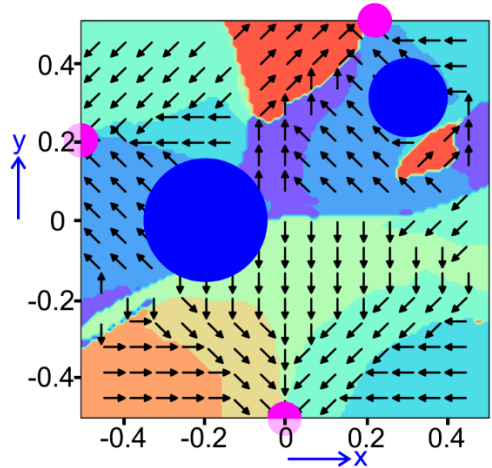

**Figure 8.** Optimal distribution of the self-driven force in the case of two round obstacles and three exits. The diameter of the obstacle near the top right corner is 2m while the other obstacle has a diameter of 3 m.

As an additional test, we study a case with 3 exits and 2 obstacles, where 80 agents are initially randomly distributed in the room. We consider the initial speed of the agents to be zero and the desired speed to be 2 m/s. Multiple obstacles are commonly present in a room and it is extremely hard to obtain an explicit solution for efficient evacuation because of the possibility of highly disordered distributions of obstacles and exits. In these cases, the proposed DRL can determine the most efficient evacuation process. The DNN has 4 hidden layers with 64 neurons for each and an output layer with 8 neurons. Following a similar training procedure as with the other cases, the resulting evacuation process is shown in Fig. 8 and Video S8. We find that the algorithm predicts which trajectory agents should follow (bypassing the obstacles or heading to another exit) based on their initial position. Overall, we show that the proposed Dyna-Q approach can handle not only simpler cases with only one exit or two exit but also more complex cases with multiple exits and obstacles.

**Conclusion**

Here, we investigate a room evacuation problem using the DRL learning approach in a particle dynamics environment. We apply a DNN to approximate the Q-value function. The DNN takes the location and velocity of an agent as input and outputs 8 discrete Q-values representing the cumulative rewards for each one of the 8 possible directions. The weights in the DNN are updated by employing the Dyna-Q learning approach, which incorporates both the model-free Q learning algorithm and the model-based planning method. Upon training completion, a single agent is able to follow the optimal policy by selecting the direction that gives the maximum total reward. After one agent learns to evacuate efficiently, the trained network is transferred to multiple agents moving in a similar environment. Overall, the Dyna-Q learning approach can efficiently train agents to evacuate a room with multiple exits and obstacles. The results are in agreement with the social force model for convex obstacles and both models can produce similar evacuation paths in simple environments when they start from the same initial configuration. However, the DRL approach has a clear advantage over the original social force model without additional layers when concave obstacles are involved. Concave obstacles can sometimes act as traps, depending on their orientation, and consequently can inhibit complete room evacuation when the social force model is implemented. Our proposed DRL approach however can derive a policy that forces object avoidance and allows for complete room evacuation. Finally, we note that there are several limitations to this model which are similar to limitations of the social force model. The most important limitation is that the optimal policy is learned by considering a single agent at different initial positions and it does not consider the motion of other agents during training. Then, the resulting policy for one agent is transferred to the other agents depending on their current position. This is not a significant problem if there is only one exit because all agents have to evacuate from this exit, even if they have to wait. It becomes an issue when there are multiple exits and/or obstacles. Then, agents must be able to dynamically choose one of the exits during evacuation. For example if crowding at one exit increases the evacuation time, agents must be able to choose to run to another exit where evacuation is faster. Another limitation of this work is that the environment is considered static. In reality however, for example during fire, the number of available exits and the number and location of obstacles along with the availability of the configuration space to the agents can change with time. In future work, we plan to modify the proposed training strategy by introducing space with a dynamic configuration where the environment can change with time and an agent will be able to consider the other agents as part of its environment.

## Acknowledgments

Y.Z., Z.C., G.L. were supported by the National Science Foundation (NSF), Mechanical and Manufacturing Innovation (CMMI) Career Award 1351363.

**SUPPLEMENTAL MATERIAL**

**Glossary:**

Reinforcement learning (RL) – A type of machine learning technique where an agent learns to obtain maximum reward by interacting with its environment.

Deep reinforcement learning (DRL) – Implementation of deep neural network in the reinforcement learning algorithm.

Model-based reinforcement learning – A reinforcement learning method based on a model of the environment.

Model-free reinforcement learning – A reinforcement learning method without a model of the environment.

Q-learning – A model-free reinforcement learning algorithm for estimating the optimal action value functions via a one-step temporal difference.

Dyna-Q learning – A reinforcement learning technique that incorporates both the model-based reinforcement learning and the model-free Q-learning method.

Transfer learning – A technique that is widely used in reinforcement learning applications. It accelerates the learning speed by training the agent in an environment and transferring the learned network parameters to a similar environment.

Experience replay – A technique used in deep reinforcement learning to avoid a strong correlation within a sequence of observations and reduce the variance between updates.

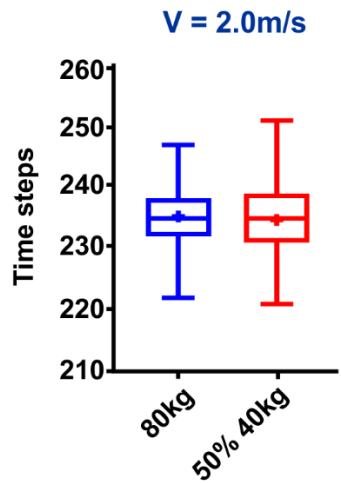

**Figure S1.** Box-and-whisker plots of the number of time steps required for 80 agents to evacuate a 10 m × 10 m room with one door and without obstacles using the Dyna-Q algorithm. The agents have zero initial speed and 2 m/s desired speed. We ran 100 cases with agents having a mass of 80 Kg starting from random initial positions and 100 cases of agents, where 50% of them have a mass of 80 kg and the rest a mass of 40 kg, starting from different random initial positions. The median value is denoted as the line in the center of the box and the mean value as a cross. We found that there is no significant difference between the median values of the two results (Mann-Whitey U test is $0.71 > 0.05$).

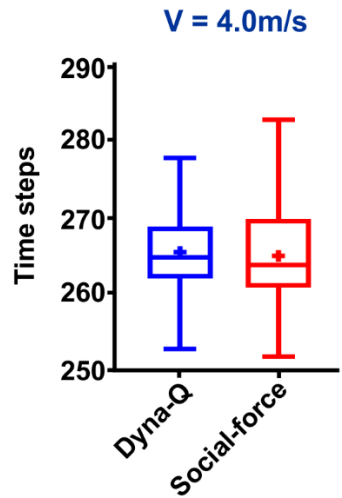

**Figure S2.** Box-and-whisker plots of the number of time steps required for 80 pedestrians to evacuate a 10 m × 10 m room with one door and without obstacles. The agents have zero initial speed and 4 m/s desired speed. We ran 100 cases of agents starting from random initial positions employing the Dyna-Q model and 100 cases of agents starting from different random initial positions employing the social force model. The median value is denoted as the line in the center of the box and the mean value as a cross. We found that there is no significant difference between the median values of the required number of steps for complete evacuation using the social force model compared to the Dyna-Q algorithm (Mann-Whitney U-test, p-value = 0.30 > 0.05).

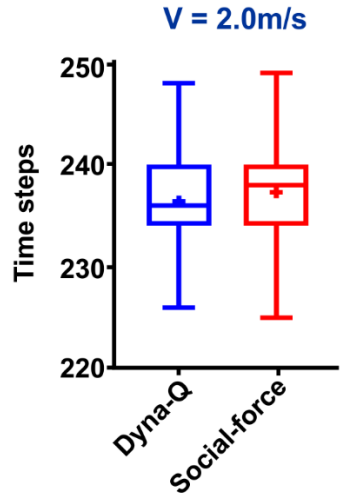

**Figure S3.** Box-and-whisker plots of the number of time steps required for 80 pedestrians to evacuate a 10 m × 10 m room with one door and without obstacles. The agents have zero initial speed and 2 m/s desired speed. We ran 100 cases of agents starting from random initial positions employing the Dyna-Q model and 100 cases of agents starting from the same random initial positions employing the social force model. The median value is denoted as the line in the center of the box and the mean value as a cross. We found that there is no significant difference between the median values of the required number of steps for complete evacuation using the social force model compared to the Dyna-Q algorithm (Wilcoxon signed rank test, p-value = 0.21 > 0.05).

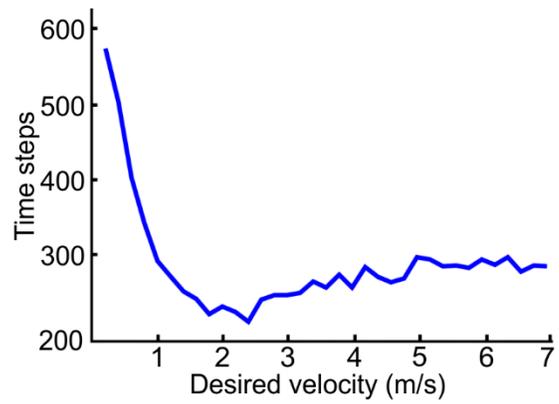

**Figure S4.** Number of time steps required for 80 pedestrians to evacuate a 10 m × 10 m room with one door and without obstacles. The agents have zero initial speed and desired speed varied from 0.2 m/s to 7 m/s. The minimum number of steps occurs at approximately 2 m/s.

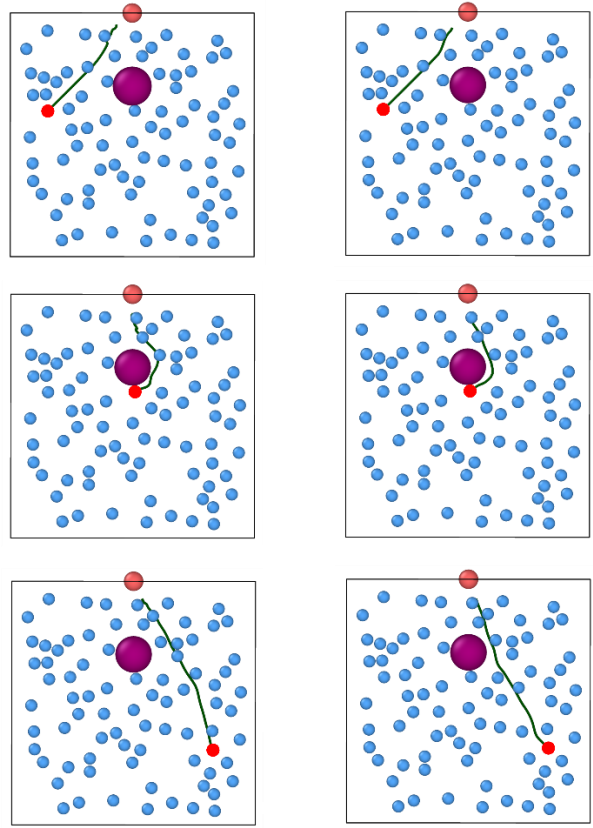

**Figure S5.** Comparison between evacuation trajectories of three, out of 80, random agents that follow the Dyna-Q model (left column) and the same three agents that follow the social force model (right column) during evacuation of a 10 m × 10 m room with one door and with a circular obstacle located close to the exit. In all three cases, the trajectories are similar.

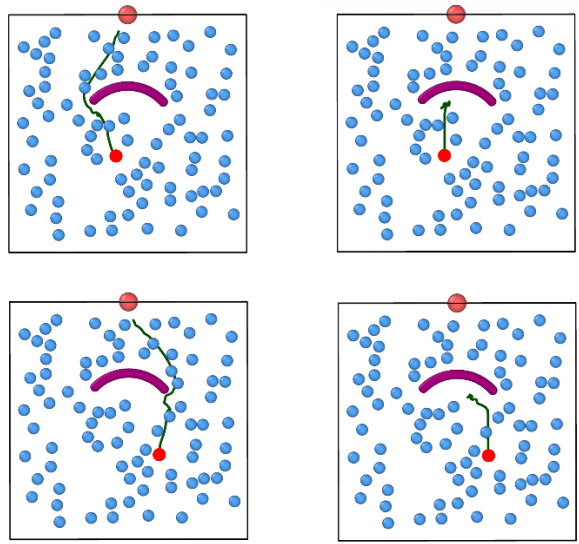

**Figure S6.** Comparison between evacuation trajectories of two, out of 80, random agents that follow the Dyna-Q model (left column) and the same three agents that follow the social force model (right column) during evacuation of a 10 m × 10 m room with one door and with a concave obstacle located close to the exit. We find that agents following the social force model can be trapped by the concave obstacle. In the DRL approach on the other hand, agents are able to identify, without additional assumptions, an efficient room evacuation policy by learning to avoid the obstacle.

**Captions for the supplemental videos:**

**Video S1:** 80 agents evacuate a 10 m × 10 m room with one exit and without obstacles following the Dyna-Q model.

**Video S2:** 80 agents following the social force model cannot fully evacuate a 10 m × 10 m room with one exit and with one concave obstacle.

**Video S3:** 80 agents following the Dyna-Q model successfully evacuate a 10 m × 10 m room with one exit and with one concave obstacle.

**Video S4:** 80 agents evacuate a 10 m × 10 m room with two exits and without obstacles following the Dyna-Q model.

**Video S5:** 80 agents evacuate a 10 m × 10 m room with four exits and without obstacles following the Dyna-Q model.

**Video S6:** 80 agents evacuate a 10 m × 10 m room with two exits and with one circular obstacle in the middle of the room following the Dyna-Q model.

**Video S7:** 80 agents evacuate a 10 m × 10 m room with two exits and with one circular obstacle located closer to the upper exit following the Dyna-Q model.

**Video S8:** 80 agents evacuate a 10 m × 10 m room with three exits and with two circular obstacles located closer to the upper exit following the Dyna-Q model.